\documentclass{bmvc2k}

\usepackage{geometry}
\usepackage{multirow}
\usepackage{makecell}
\usepackage{colortbl}

\title{Sparse Multi-Object Render-and-Compare}

\addauthor{Florian Langer}{fml35@cam.ac.uk}{1}
\addauthor{Ignas Budvytis}{ib255@cam.ac.uk}{1}
\addauthor{Roberto Cipolla}{rc10001@cam.ac.uk}{1}
\addinstitution{
 Department of Engineering\\
 University of Cambridge\\
 Cambridge, UK
}

\runninghead{Langer et al.}{Sparse Multi-Object Render-and-Compare}

\usepackage{array}
\newcommand{\PreserveBackslash}[1]{\let\temp=\\#1\let\\=\temp}
\newcolumntype{C}[1]{>{\PreserveBackslash\centering}p{#1}}
\newcolumntype{R}[1]{>{\PreserveBackslash\raggedleft}p{#1}}
\newcolumntype{L}[1]{>{\PreserveBackslash\raggedright}p{#1}}

\begin{document}

\maketitle

\begin{abstract}
Reconstructing 3D shape and pose of static objects from a single image is an essential task for various industries, including robotics, augmented reality, and digital content creation.
This can be done by directly predicting 3D shape in various representations \cite{meshrcnn,sdf,compressive_coding} or by retrieving CAD models from a database and predicting their alignments \cite{mask2cad,patch2cad,geometric_correspondence_fields,roca,sparc}. Directly predicting 3D shapes often produces unrealistic, overly smoothed or tessellated shapes \cite{meshrcnn,pixel2mesh,total_3D_understanding}. Retrieving CAD models ensures realistic shapes but requires robust and accurate alignment. Learning to directly predict CAD model poses from image features is challenging and inaccurate \cite{mask2cad,patch2cad}. Works, such as ROCA \cite{roca}, compute poses from predicted normalised object coordinates which can be more accurate but are susceptible to systematic failure. SPARC \cite{sparc} demonstrates that following a ``render-and-compare'' approach where a network iteratively improves upon its own predictions achieves accurate alignments. Nevertheless, it performs individual CAD alignment for every object detected in an image. This approach is slow  when applied to many objects as the time complexity increases linearly with the number of objects and can not learn inter-object relations. Introducing a new network architecture Multi-SPARC we learn to perform CAD model alignments for multiple detected objects jointly.
Compared to other single-view methods we achieve state-of-the-art performance on the challenging real-world dataset ScanNet \cite{scannet}.  By improving the instance alignment accuracy from $31.8\%$ \cite{roca} to $40.3\%$ we perform similar to state-of-the-art multi-view methods \cite{vid2cad}.

\end{abstract}

\section{Introduction}

Approaches to reconstructing 3D scenes from an image can be broadly split up into direct shape prediction \cite{meshrcnn,pixel2mesh,total_3D_understanding} as well as retrieval-based methods \cite{mask2cad,patch2cad,geometric_correspondence_fields,roca}. The issue with the former is that they struggle to reconstruct high quality shapes. Retrieval-based approaches on the other hand often have difficulty in accurately aligning CAD models to an image. Some existing works \cite{mask2cad,patch2cad} directly regress CAD model poses from image features. Whilst being simple such methods are often inaccurate. Other methods, such as ROCA \cite{roca}, predict dense 2D to 3D correspondences and use these correspondences for computing object poses. While such approaches allow for more accurate pose estimates, the predicted correspondences are often systematically shifted, leading to a constant offset in the alignment. 
\begin{figure*}[t]
    \centering
    \includegraphics[width=1.0\linewidth]{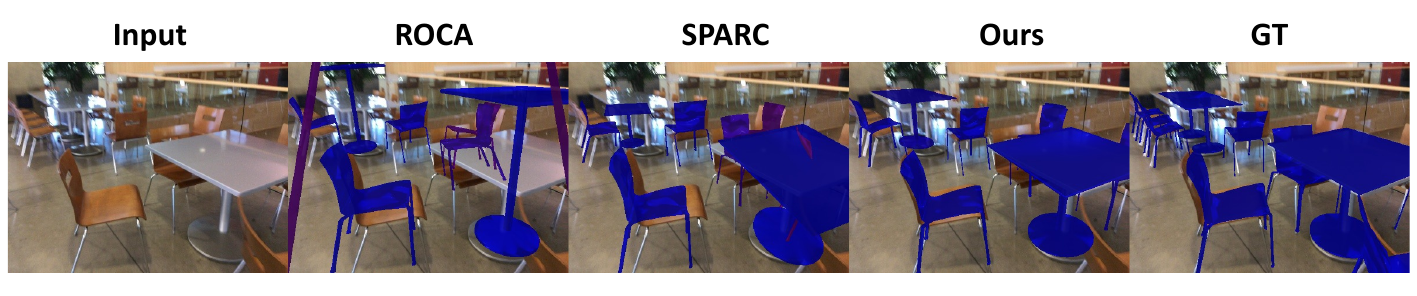}
    \vspace{-0.3cm}
    \caption{\textbf{State-of-the-art methods for CAD model alignment from a single image.} Using normalised object coordinates ROCA's \cite{roca} alignments suffer from constant offsets. SPARC \cite{sparc} produces more accurate alignments, but predicts CAD model poses individually which is slow and leads to worse predictions. In our method CAD model alignments are predicted jointly which is faster and more accurate.}
    \vspace{-0.3cm}
    \label{fig_intro}
\end{figure*}
Recent work \cite{sparc} demonstrates that an iterative, render-and-compare approach is more accurate and robust than relying on normalised object coordinates. However, \cite{sparc} perform CAD model alignment individually for every detected object which is slow at test time and can not model inter-object relations. We introduce a render-and-compare approach to deal with multiple CAD models simultaneously. For this purpose we predict bounding boxes, surface normals, depth and segmentation masks for a given input image. For every detected bounding box we initialise a CAD model in some initial pose and reproject points and surface normals sampled from the CAD model into the image plane. This information in combination with sparse information about the depth, surface normals, segmentation masks and RGB is used as the input to a Perceiver-based \cite{perceiver} alignment network which predicts pose updates for all CAD models jointly.\\
We demonstrate that learning pose alignments jointly and pre-training our network on a large number of randomly sampled synthetic scenes leads to state-of-the-art-performance on the real-world dataset ScanNet \cite{scannet}. Another important observation is that our network benefits from imposing some structure on the latent space. In addition to learning pose alignments we learn classification scores indicating whether the current alignment is accurate or not. We show that we can use these classification scores to select the best alignment from different initialisations. Our system improves the instance alignment accuracy on ScanNet \cite{scannet} from $31.8\%$ \cite{roca} to $40.3\%$. 
In summary our contributions include:
\begin{itemize}
    \vspace{-0.2cm}
    \item A novel render-and-compare approach which jointly predicts CAD model alignments for multiple CAD models simultaneously;
    \vspace{-0.2cm}
    \item A demonstration that synthetic pre-training on a large number of synthetic scenes achieves state-of-the-art performance on the challenging real-world dataset ScanNet \cite{scannet}.
    \vspace{-0.2cm}
    \item A well calibrated classification score that can be used for selecting CAD model poses from different initialisations and other tasks.
\end{itemize}

\section{Related Work}
Aligning CAD models to images is a form of 3D reconstruction. While there exist a large number of works that perform 3D reconstruction by directly predicting shapes in various representations \cite{meshrcnn,pixel2mesh,binary_planes,pointclouds,point_set,polytopes,primitives}, this section will focus on works that, like ours, perform 3D reconstruction by retrieving CAD models and aligning them to images. Those works can be split along two meaningful axes: Whether they are single-shot predictions or perform iterative render-and-compare, or whether they predict object poses individually or for multiple CAD models jointly.\\
\textbf{Single-shot alignments vs. iterative procedures}.
Mask2CAD \cite{mask2cad} and Patch2CAD \cite{patch2cad} directly predict CAD model poses by simply regressing the 6-DoF pose with a convolutional network. While this approach is very simple and fast it is not very accurate and performs poorly for unseen objects. \cite{langer_leveraging_shape} demonstrate more accurate alignments by establishing sparse 2D-3D correspondences between RGB images and rendered CAD model and use these constraints to find the pose that maximizes the silhouette overlap with an instance segmentation prediction. ROCA \cite{roca} demonstrate a more robust method by leveraging predicted depth to lift dense 2D-3D correspondences into 3D and directly optimizing for the pose that minimizes the 3D correspondence error. 
In contrast to these works stand approaches that iteratively update a CAD model pose. These works include \cite{Im2CAD} and \cite{geometric_correspondence_fields} which learn a comparison function between the original image and the rendered CAD model. Both works maximise the learned similarity function at test time using gradient descent requiring 250 to 1000 update steps with run-times of 4 minutes and 36 seconds respectively. SPARC \cite{sparc} demonstrate that render-and-compare can be harnessed more efficiently by directly learning to predict pose updates which proves to be a lot faster (2 seconds) and more robust to poor initialization. 
Our method works similar to SPARC \cite{sparc} but we demonstrate how to apply render-and-compare to multiple objects simultaneously.\\
\textbf{Single-object vs. multi-object.}
\cite{roca,mask2cad,patch2cad,sparc,geometric_correspondence_fields} all predict alignments for every CAD model individually. While \cite{roca,mask2cad,patch2cad} are still fast as they use the same encoder for making predictions for multiple CAD models,  \cite{sparc,geometric_correspondence_fields} need to perform render-and-compare separately for every object which is slow at test time as the time increases linearly with the number of objects in the scene. This can be very slow for scenes with many objects. Independent of the speed all of these methods fail to model inter-object relations which are valuable when attempting to predict accurate CAD alignments.\\
Methods like \cite{scenecad,holistic_3d_from_image,total_3D_understanding} explicitly model inter-object relations demonstrating that these can contain valuable information for the alignment. \cite{scenecad,holistic_3d_from_image,total_3D_understanding} model object relations with a graph where nodes represent objects and edges represent their relations with each other. In comparison we allow our network to learn object relations by imposing less structure by having a dense latent space where information from different objects can attend to information regarding its own alignment and the alignment of other objects through attention.
\section{Method}

\begin{figure*}[t]
    \centering
    \includegraphics[width=1.0\linewidth]{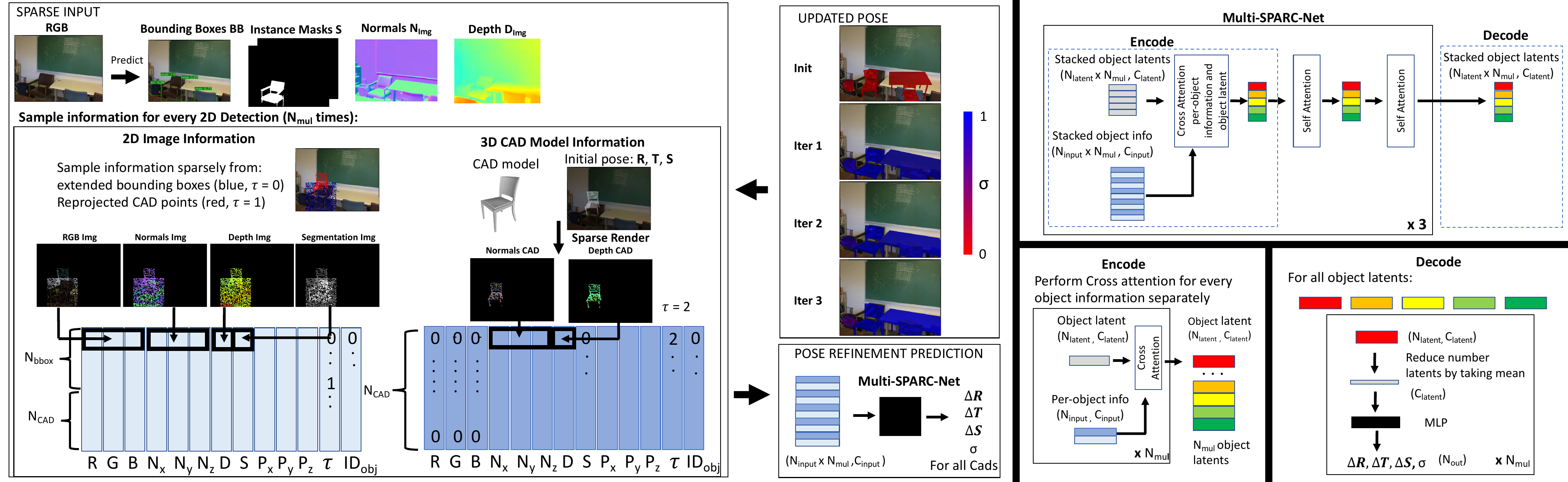}
    \vspace{-0.3cm}
    \caption{\textbf{Method}: Left side: For all 2D detections we sample the RGB values (RGB), surface normals (N), depth values (D) and instance segmentation mask values (S) from inside the detected bounding boxes and for pixel bearing $(P_x,P_y,P_z)$ onto which a 3D CAD point is reprojected. CAD model information is encoded by reprojecting 3D points and surface normals into the image plane. Right side: Using Multi-SPARC-Net we encode information for each alignment separately into a latent space using cross-attention. Repeating blocks of separate cross-attention followed by self-attention layers three times we decode from each part of the latent space separately to predict pose updates $\Delta \textbf{R}$, $\Delta \textbf{T}$ and $\Delta \textbf{S}$ as well as a classification score $\sigma$. Pose updates are used to iteratively refine CAD model poses and $\sigma$ is used for choosing the best alignment from different rotation initialisations (see Fig. \ref{fig_confidence}a).}
    \vspace{-0.2cm}
    \label{fig_method}
\end{figure*}
In this section we describe the three key steps of our method: (i) 2D object detection, instance segmentation as well as surface normal and depth estimation (Sec. \ref{sec_method_1}), (ii) sparse input generation (Sec. \ref{sec_method_2}) and (iii) pose update predictions (Sec. \ref{sec_method_3}) where we iteratively repeat steps (ii) and (iii). Sec. \ref{sec_method_4} explains the synthetic pre-training we used.

\subsection{Object Detection, Instance Segmentation, Normal and Depth Prediction}
\label{sec_method_1}
As a first step we perform 2D object detections by predicting a set of bounding boxes (BB) and object classes (see Fig. \ref{fig_method}) using Mask-RCNN \cite{maskrcnn}. We use the same bounding boxes, object classes and CAD model retrievals as ROCA \cite{roca}, although any other method could be employed as well. Additionally, we use instance segmentation predictions (S) from \cite{segany} prompted with the detected bounding boxes.
For estimating surface normals (N) and depth values (D) we follow the same training procedure as \cite{sparc}. We employ a lightweight convolutional encoder-decoder architecture from \cite{GB-mono-2018-densedepth}. The training losses are consistent with state-of-the-art works for surface normal estimation \cite{GB-SNfromRGB_21_BAE} and for depth estimation \cite{Bae_2022_CVPR_depth}. We use ground truth surface normals provided by \cite{GB-SNfromRGB_19_FrameNet} and ground truth depth from ScanNet \cite{scannet} (for more details see the Supp. Mat.). When training the surface normal and depth estimation network, we respect the train and test split used in our evaluation.

\subsection{Generating Sparse Inputs}
\label{sec_method_2}
Rather than processing full images we sample sparse image information as vectors through different image channels \cite{sparc}. We sample the location of those vectors from two regions, inside the detected bounding boxes (blue points in Fig. \ref{fig_method}) and from pixels onto which 3D CAD model points were reprojected (red points). The different input channels include their color values (RGB), surface normal (N) and depth estimates (D) as well as their instance segmentation mask value (S). We append to those vectors the corresponding pixel bearing $(P_x,P_y,P_z)$ (to provide information on the location of the sampled values), a token $\tau$ corresponding to the type of input ($\tau = 0$ for bounding box, $\tau = 1$ for reprojected points) and the ID of the detection. For a single detection all vectors are stacked to make up the light blue block of shape $(N_{bbox} + N_{CAD},C_{input})$ in Fig. \ref{fig_method}. We encode the 3D CAD model information of shape $(N_{CAD},C_{input})$ (dark blue block) in a similar way by sampling 3D points and corresponding surface normals from the CAD model in the current pose $\textbf{R},\textbf{T},\textbf{S}$. When reprojecting those points into the image plane we can compute the locations of the corresponding pixel bearings and the values of their surface normal and depth. Values for the color channels (RGB) and instance segmentation (S) are filled with zeros. For the region channel we add $\tau = 2$ and also include the detection ID. 
Together, both blocks of information make up all the information for a given detection which is encoded separately into the latent space. This information is sampled for all detections up to a maximum number of $\text{N}_{mul}$ detections. If there are fewer detections than $\text{N}_{mul}$ inputs are padded with zeros. If there are more detections, they are split up into multiple forward passes.

\subsection{Pose Update Predictions}
\label{sec_method_3}
This subsection provides details on the network architecture, pose parameterisation, loss function and iterative refinement procedure.\\
\textbf{Network Architecture.}
Our network architecture is built on a Perceiver network \cite{perceiver} with one small difference. Rather, than encoding all input information of the different detections jointly we found it beneficial to encode them separately using a shared cross-attention layer ($[N_{input},C_{input}],[ N_{latent},C_{latent}] \rightarrow [N_{latent},C_{latent}])$ (see Fig. \ref{fig_method} right side). We concatenate all encodings and apply two layers of self-attention ($[N_{mul} \cdot N_{latent},C_{latent}] \rightarrow [N_{mul} \cdot N_{latent},C_{latent}]$) which allows for processing information relevant to the alignment and for sharing information between the different alignments. This block of per-object cross-attention followed by two layers of self attention is repeated three times. At the decoding stage we again decode from the relevant portion of the latent space for each detection separately. For this we reduce the $[N_{\text{Latent}},C_\text{Latent}]$ latent space for each object to an $[C_\text{Latent}]$ embedding by taking the mean over the first dimension. We map this to the desired number of output parameters $N_\text{out} = 11$ using an MLP. The same MLP is applied to the different portions of the latent space to produce pose updates for every detection.\\
\textbf{Pose Parameterisation.}
The outputs are the updates to the current pose $(\textbf{T},\textbf{R},\textbf{S})$. They consist of a translation update $\Delta \textbf{T}$, a rotation update $\Delta \textbf{R}$ and a scale update $\Delta \textbf{S}$ as well as a classification score $\sigma$ indicating whether the starting pose was already an accurate alignment or not. 
We parameterise $\textbf{T}$ with polar coordinates ($d,\phi,\theta$) where $d$ is the distance from the camera center and $\phi$ and $\theta$ parameterise a vector on the unit sphere. The updated translation $\textbf{T}'$ is given by $ \textbf{T}' = ( d \cdot \Delta d,\phi + \Delta \phi,\theta + \Delta \theta)$. Rotation is parameterised using quarternions which are transformed to a rotation matrix before making the rotation update $\textbf{R}' = \textbf{R}
 \cdot \Delta \textbf{R}$. Finally, $\textbf{S}$ is parameterised by three axis-aligned scaling parameters and $\textbf{S}' = \textbf{S} \cdot \Delta \textbf{S}$.
The updates for scale and the distance parameter $d$ are multiplicative rather than additive. This is to ensure that the learned updates are decoupled from each other as much as possible. An additive scale update will produce different effects depending on whether the object is close and small or far away and large. In contrast, a multiplicative scale update will produce the same result. We ensure that the predicted updates are positive by applying a sigmoid function to the predicted values. Choosing polar coordinates was again motivated by the intuition that decoupled pose updates are easier to learn than coupled ones. While for euclidean coordinates a given $X$ prediction will have a very different effect if the object is close and small or far and large, predicting updates for $\phi$ and $\theta$ will have the same effect regardless of the distance.\\
\textbf{Loss function.} Our loss function is comprised of two components, one for learning the CAD model alignments and one for learning the pose classifications. For learning the alignments we introduce a loss function that unifies learning translation, rotation and scale, and does not require any hyper-parameter tuning for weighing the relative strengths of different components. Our loss is simply given by the L1 distance of $N_{\text{loss}}$ points $\textbf{P}$ sampled from the CAD model in the ground truth pose $(\textbf{T}_{\text{GT}},\textbf{R}_\text{GT},\textbf{S}_\text{GT})$ to the CAD model under the predicted pose $(\textbf{T}',\textbf{R}',\textbf{S}')$, $L_{\text{align}} = \sum_{i=1}^{N_{\text{loss}}} | F' ( \textbf{P}_i ) - F_{GT} ( \textbf{P}_i )|$, where $F'$ and $F_{GT}$ denote the affine transformations when applying $\textbf{S}'$,$\textbf{R}'$ and $\textbf{T}'$ or $\textbf{S}_{GT}$,$\textbf{R}_{GT}$ and $\textbf{T}_{GT}$ respectively. In general, poses are initialised from a large range of translations, rotations and scale to ensure that at test time the network is robust to poor detections. Consistent with previous work \cite{mask2cad,sparc}, we find that it is difficult to learn rotation updates over the entire rotation space. We therefore constrain initialisations to be within an azimuthal angle of $\pm 45^\circ$ of $\textbf{R}_\text{GT}$. At test time we initialise from $0^\circ$, $90^\circ$, $180^\circ$ and $270^\circ$ azimuthal angle and use the predicted pose classification $\sigma$ to identify the correct prediction.
For learning $\sigma$ we use a binary cross entropy loss. A given pose is labelled correct if its translation, rotation and scale are within $20$ cm, $20^\circ$ and 20\% respectively, $L_{\text{classifier}} = L_{\text{BCE}} (\sigma,\sigma_{\text{GT}})$. Therefore the total loss is given by $L_\text{total} = L_{\text{align}} + L_{\text{classifier}}$. In order to balance the training of the pose classifier we sample separate training poses which are different from the ones used for learning the pose updates (see the Supp. Mat.).\\
\textbf{Iterative Refinements.}
After a given prediction at train time the next initial poses will be the updated poses based on the networks predictions. This ensures that the network learns to predict pose updates for realistic poses that it is likely to encounter at test time. After repeating this 3 times a new batch of images is initialised with objects sampled in random poses. At test time pose updates are predicted for all objects in the image which are initialised from 4 different azimuthal angles rotated $90^\circ$ with respect to each other (Fig. \ref{fig_method} shows just one such initialisation). For each initialisation three pose updates are predicted and in a fourth iteration their classification score $\sigma$ is determined. For each detection the pose with the highest classification score is returned as the final prediction (see Fig. \ref{fig_confidence}a).

\subsection{Synthetic Pre-training}
\label{sec_method_4}
For the synthetic pre-training we sample random objects from 3D-Future \cite{3d_future} in random poses and render them on-the-fly with PyTorch3D \cite{pytorch3d}. We use CAD models from 3D-Future as opposed to the CAD models from ShapeNet \cite{chang2015shapenet} used for our main training and evaluation as many ShapeNet models contain holes or are poorly meshed leading to artifacts when rendering surface normals. For more details see the Supp. Mat.
\section{Experimental Setup}
\begin{figure*}[t]
    \centering
    \includegraphics[width=1.0\linewidth]{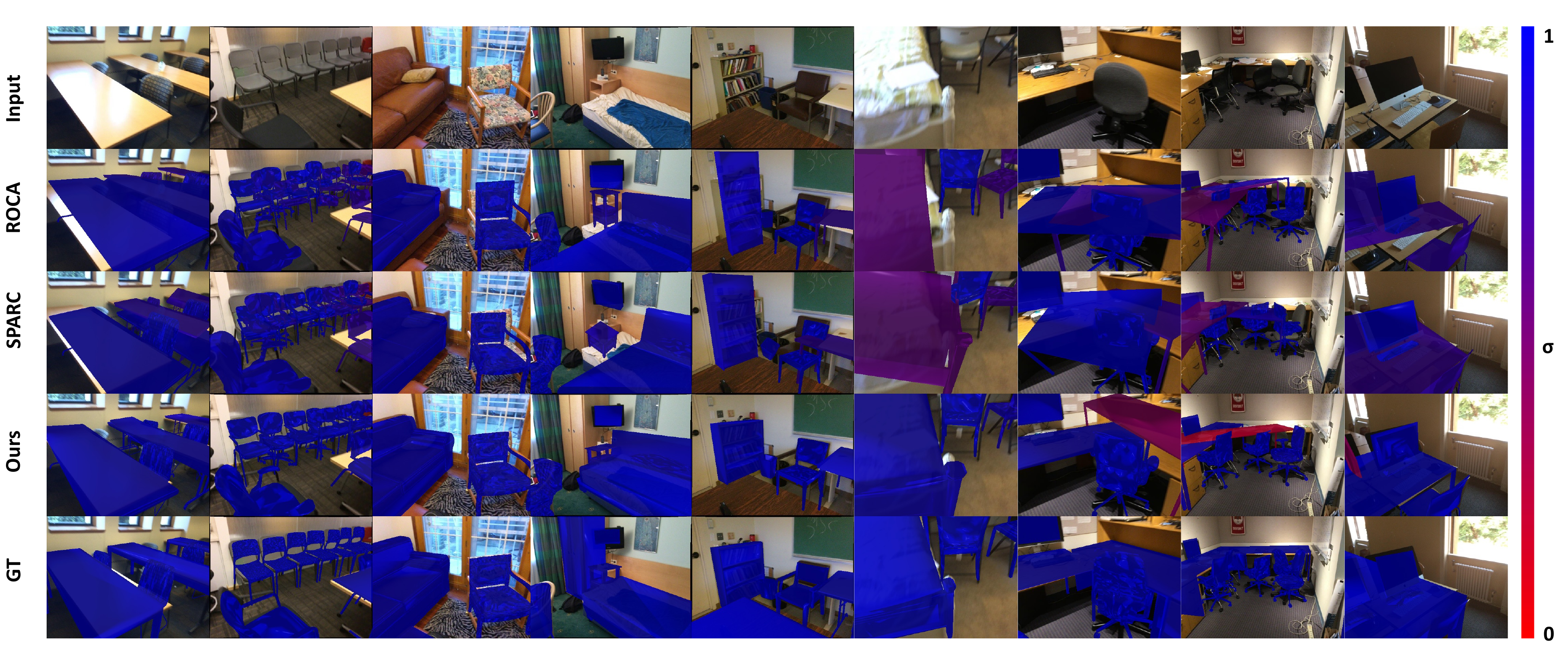}
    \vspace{-0.3cm}
    \caption{\textbf{Qualitative comparison.} Particularly for multiple objects close to each other our alignments are more accurate than existing methods (column 1 - 5). Due to the synthetic pre-training, our network can even work from very challenging viewpoints (column 6). Furthermore, our learned 3D classification score allows the network to identify potentially bad alignments (column 7 - 8). Our network struggles to correctly classify display orientations leading to poor performance on that class (column 9).}
    \vspace{-0.3cm}
    \label{fig_qualitative}
\end{figure*}
This section provides a concise overview of the dataset employed in training and testing, along with an explanation of the evaluation metrics and the selected hyperparameters.\\
\textbf{ScanNet dataset.}
Following the approach of \cite{total_3D_understanding,mask2cad,patch2cad,roca,sparc}, we use the ScanNet25k image dataset \cite{scannet} for training and testing, which includes CAD model annotations provided by \cite{scan2cad}. This dataset comprises 20,000 training images from 1,200 training scenes and 5,000 test images from 300 distinct test scenes. Our method is trained and tested on the top 9 categories with the highest number of CAD annotations covering over 2,500 unique shapes.\\
\textbf{Evaluation metrics.}
For our main evaluation we follow the original evaluation protocol established by Scan2CAD \cite{scan2cad} which evaluates CAD model alignments on a per-scene basis. We convert predicted CAD model poses into ScanNet \cite{scannet} world coordinates and, similar to \cite{roca,sparc}, apply 3D non-maximum suppression to remove multiple detections of identical objects from different images. For the evaluation, a CAD model prediction is deemed correct if the object class prediction is correct, the translation error is less than 20 cm, the rotation error is less than 20°, and the scale error is below 20\%. We report the percentage of correct alignments for each class individually as well as the overall instance alignment accuracy for all predictions.\\
In addition to the per-scene alignments we evaluate per-image alignments. For this purpose we reproject CAD models in GT poses into the individual camera frames. Note that for each camera frame only GT CAD models whose center is reprojected into the camera view are considered. For every predicted CAD alignment we find the associated GT CAD model by computing the IoU of the 2D bounding boxes and finding that GT CAD model of the same category with maximum IoU. In order to avoid penalising for objects that are not visible due to occlusion we only consider GT objects for which at least 50\% of pixel have the rendered depth value within 30 cm of the GT sensor depth value.
Similar to the per-scene metric we evaluate the alignment accuracy by computing the percentage of predictions whose errors for rotation, translation and scale are within the same thresholds as above.
Additionally we compute $\text{AP}^{\text{mesh}}$ introduced by \cite{meshrcnn}. It is defined as the mean area under the per-category precision-recall curves for $\text{F}^\rho$ at different thresholds. The $\text{F}^\rho$
score is the harmonic mean of the fraction of points sampled from the predicted aligned CAD model that are within $\rho$
of a point sampled from the GT aligned CAD model and the fraction of points sampled from the GT CAD model within $\rho$
of a point sampled from the predicted CAD model. We evaluate AP50, which considers a prediction to be correct if $\text{F}^\rho > 0.5$, as well as AP mean which takes the average across the ten AP scores AP50, AP55,...,AP95 sampled in regular intervals.\\
\textbf{Hyperparameters.}
For our inputs we sample $N_{bbox} = 2000$ pixels inside the predicted bounding box which is uniformly extended by 10\% and use $N_{CAD} = 500$ points from the CAD model. $N_{input} = (N_{bbox} + 2N_{CAD})$ and $C_{input} = 13$.
We set the number of latents $N_{latent} = 80$ where each latent has   $C_{latent} = 256$ channels. We choose $N_{mul} = 5$ which means that a maximum of 5 CAD models are processed jointly. If an image contains more than 5 detections the detections are split into multiple blocks. We show in the Supp. Mat. that we achieve similar results with larger numbers of $N_{mul}$. 
We use batches of 20 images and use the Lamb optimisier \cite{lamb} with learning rate set to $0.001$. We sample $N_{loss} = 1000$ points for computing the loss.
Our model is pretrained on 10 M rendered images containing between 1 and 4 CAD models in random poses.\\
\textbf{Implementation Details.}
All code is implemented in PyTorch. Pre-training our main model takes 6 days on a single TitanXp. Finetuning on ScanNet25k for 500 epochs takes 2 days. 
\section{Results}
\begin{table}[t]
    \centering
    \resizebox{\textwidth}{!}{
    \begin{tabular}{c|c|ccccccccc|cc|c}
     &Method & bathtub & bed & bin & bkshlf & cabinet & chair & display & sofa & table & \cellcolor[gray]{0.8} \textbf{class} & \cellcolor[gray]{0.8} \textbf{instance} & time [ms] \\ [0.5ex] 
     \hline
     & Number of Instances \# & 120 & 70 & 232 & 212 & 260 & 1093 & 191 & 113 & 553 & \cellcolor[gray]{0.8} 9 & \cellcolor[gray]{0.8} 2844 & - \\
     \hline
      \multicolumn{14}{c}{\textbf{Ablation experiments}}\\
     \hline
     \multirow{2}{*}{Joint Encoding and Decoding}
     & separate encoding - joint decoding & 17.5 & 21.4 & 26.7 & 9.9 & 15.0 & 45.7 & 3.1 & 29.2 & 17.5 & \cellcolor[gray]{0.8} 20.7 & \cellcolor[gray]{0.8} 27.9 & 864 \\
     & joint encoding - separate decoding & 18.3 & 34.3 & 36.6 & 12.7 & 14.2 & 52.8 & 4.7 & 25.7 & 17.5 & \cellcolor[gray]{0.8} 24.1 & \cellcolor[gray]{0.8} 31.9 & 656 \\
     \hline
     \multirow{4}{*}{\makecell{Single vs Multi \\ Pre-training vs. No Pre-trainig}}
    & single-object - no pre-training & 20.8 & 24.3 & 39.2 & 12.7 & 22.7 & 57.5 & 2.1 & 24.8 & 19.2 & \cellcolor[gray]{0.8} 24.8 & \cellcolor[gray]{0.8} 34.6 & 2320\\
    & \cellcolor{yellow} multi-object - no pre-training & \cellcolor{yellow} 20.0 & \cellcolor{yellow} 28.6 & \cellcolor{yellow} 40.1 & \cellcolor{yellow} 13.7 & \cellcolor{yellow} 20.4 & \cellcolor{yellow} 59.9 & \cellcolor{yellow} 0.5 & \cellcolor{yellow} 36.3 & \cellcolor{yellow} 23.0 & \cellcolor{yellow} 26.9 & \cellcolor{yellow} 36.7 & \cellcolor{yellow} 864 \\
    & single-object - pre-training & 28.3 & 35.7 & 36.6 & 20.3 & 23.1 & 61.4 & 4.8 & 37.2 & 23.5 & \cellcolor[gray]{0.8} 30.2 & \cellcolor[gray]{0.8} 38.7 & 2320\\
    & multi-object - pre-training & 25.8 & 34.3 & 44.8 & 17.0 & 19.2 & 64.8 & 5.8 & 35.4 & 25.5 & \cellcolor[gray]{0.8} 30.3 & \cellcolor[gray]{0.8} 40.3 & 864\\
    \hline
    \multirow{2}{*}{Sparser and Faster} & $N_{bbox} = 200$, $N_{CAD} = 200$ - joint encoding& 25.0 & 34.3 & 33.6 & 14.2 & 17.7 & 56.6 & 2.6 & 35.4 & 21.2 & \cellcolor[gray]{0.8} 26.7 & \cellcolor[gray]{0.8} 34.8 & 480 \\
     & $N_{bbox} = 50$, $N_{CAD} = 50$ - joint encoding & 14.2 & 25.7 & 31.0 & 9.9 & 18.1 & 55.0 & 7.3 & 29.2 & 22.8 & \cellcolor[gray]{0.8} 23.7 & \cellcolor[gray]{0.8} 33.4 & \textbf{448} \\
     \hline
     \multirow{2}{*}{Learned Classification Score}
     & 2D confidence & 27.5 & 31.4 & 45.3 & 16.0 & 20.4 & 60.6 & 5.8 & 38.9 & 25.1 & \cellcolor[gray]{0.8} 30.1 & \cellcolor[gray]{0.8} 38.8 & 816 \\
     & 3D classification & 25.8 & 34.3 & 44.8 & 17.0 & 19.2 & 64.8 & 5.8 & 35.4 & 25.5 & \cellcolor[gray]{0.8} 30.3 & \cellcolor[gray]{0.8} 40.3 & 864\\
    \hline
    \multicolumn{14}{c}{\textbf{Comparison to other methods - per-scene evaluation}}\\
    \hline
     \multirow{4}{*}{Single-view} & Total3D-ODN \cite{total_3D_understanding}& 10.0 & 2.9 & 16.8 & 2.8 & 4.2 & 14.4 & 13.1 & 5.3 & 6.7 & \cellcolor[gray]{0.8} 8.5 & \cellcolor[gray]{0.8} 10.4 & - \\
     & Mask2CAD-b5 \cite{mask2cad} & 7.5 & 2.9 & 24.6 & 1.4 & 5.0 & 29.9 & 13.1 & 5.3 & 5.6 & \cellcolor[gray]{0.8} 10.6 & \cellcolor[gray]{0.8} 16.7 & 60\\ 
     & ROCA \cite{roca} & 20.8 & 8.6 & 26.3 & 9.0 & 13.1 & 39.9 & \textbf{24.6} & 10.6 & 12.7 & \cellcolor[gray]{0.8} 18.4 & \cellcolor[gray]{0.8} 25.0 & 53 \\
    & SPARC \cite{sparc} & \textbf{25.8} & 25.7 & 24.6 & 14.2 & \textbf{20.8} & 51.5 & 17.8 & 28.3 & 15.4 & \cellcolor[gray]{0.8} 24.9 & \cellcolor[gray]{0.8} 31.8 & 1925 \\
     & \cellcolor{orange} Ours & \cellcolor{orange} \textbf{25.8} & \cellcolor{orange} \textbf{34.3} & \cellcolor{orange} \textbf{44.8} & \cellcolor{orange} \textbf{17.0} & \cellcolor{orange} 19.2 & \cellcolor{orange} \textbf{64.8} & \cellcolor{orange} 5.8 & \cellcolor{orange} \textbf{35.4} & \cellcolor{orange} \textbf{25.5} & \cellcolor{orange} \textbf{30.3} & \cellcolor{orange} \textbf{40.3} & \cellcolor{orange} 864\\
    \hline
    Multi-view & Vid2CAD \cite{vid2cad} & 27.5 & 35.7 & 45.7 & 9.9 & 21.5 & 63.4 & 33.0 & 24.8 & 25.8 & \cellcolor[gray]{0.8} 31.9 & \cellcolor[gray]{0.8} 41.0 & 2500\\
    \hline
    \end{tabular}}
    \vspace{0.1cm}
    \caption{\textbf{Alignment Accuracy on ScanNet}     
    \cite{scannet,scan2cad} in \% for the per-scene evaluation in comparison to the state-of-the-art. Bolded numbers denote the highest accuracy for the single-view methods. Times are for reconstructing a scene containing 5 objects. The yellow row highlights the reference for comparing ablations for ``joint encoding and decoding'' as well as the ``sparser and faster'' experiments for which no pre-training was performed. The orange row are our main results.\vspace{-0.3cm}}
    \label{table_main_results}
    
\end{table}

\begin{figure*}[t]
    \centering
    \includegraphics[width=1.0\linewidth]{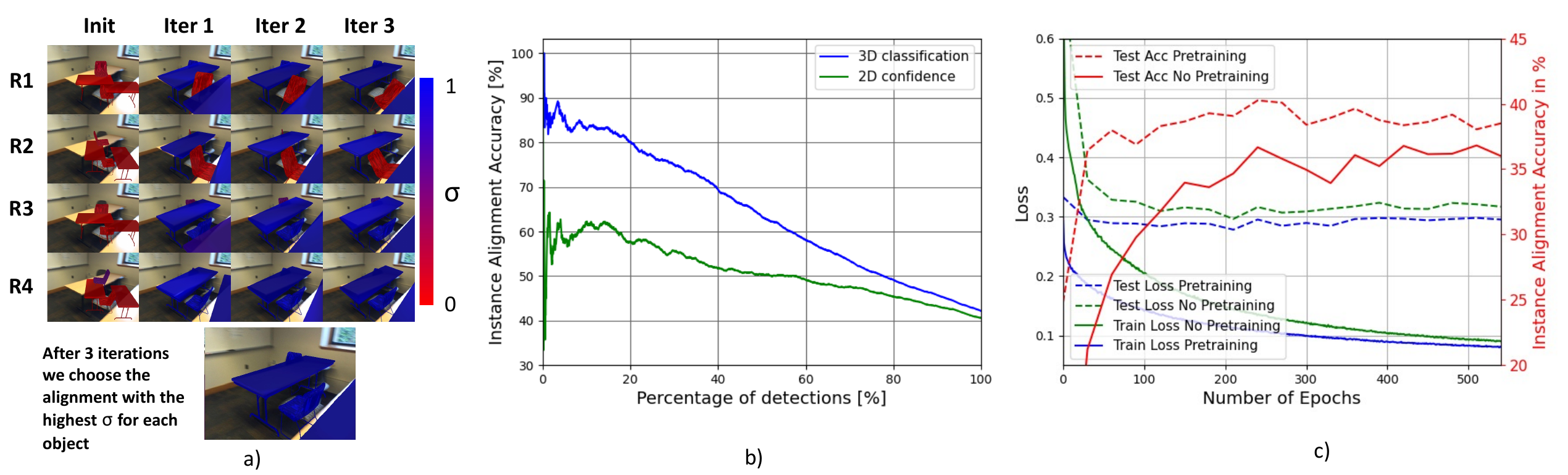}
    \vspace{-0.4cm}
    \caption{\textbf{Pose selection, calibration and loss functions.} a) We use the predicted classification score to select the final object predictions from 4 different rotation initialisations. b) The classification score is also used in the ScanNet evaluation to filter out duplicate predictions. Compared to the 2D confidence  scores (green) from \cite{roca} our 3D classification score (blue) is significantly better calibrated. c) Synthetic pre-training leads to lower losses during training and testing as well as a higher instance alignment accuracy on the test set.}
    \vspace{-0.5cm}
    \label{fig_confidence}
\end{figure*}

This section explains our qualitative and quantitative results. We first ablate major design choices in the network architecture and training procedure and subsequently compare our method to the state-of-the-art. If not stated otherwise numbers in the following refer to the overall instance alignment accuracy of all objects on ScanNet \cite{scannet}.\\
\textbf{Separate Encoding and Decoding.}
When performing multi-CAD model alignment with a transformer-based \cite{attention_all_need} architecture, naively one would simply concatenate all inputs, marking information for different alignments with different tokens, and hoping that the network will learn to regress all pose updates jointly. The first two rows in Tab. \ref{table_main_results} show results for the experiments where we perform joint decoding or joint encoding. For the former we reduce all latents
$[N_{mul} \cdot N_{latent} ,C_{latent}] \rightarrow [C_{latent}]$ by taking the mean over the first dimension and then learning an MLP to map to
 $N_{mul} \cdot N_{out}$ directly. For the latter we have one large cross attention that maps from all the concatenated inputs to all latents $([N_{mul} \cdot N_{input} ,C_{input}],[N_{mul} \cdot N_{latent} ,C_{latent}] \rightarrow [N_{mul} \cdot N_{latent} ,C_{latent}])$.
 Comparing the instance alignment accuracy $27.9\%$ and $31.9\%$ to the alignment accuracy for the multi-object results without pre-training $36.7\%$ we find that both separate encoding and separate decoding are crucial for good alignments, with separate decodings being even more important. The intuition behind this is that it is not easy for the network to learn to associate input information from different CAD models to the correct output values and encoding and decoding separately helps with this.\\
\textbf{Single vs. Multi-object and Pre-training vs. No Pre-training.}
Our experiments show that performing CAD model alignments jointly leads to slightly more accurate alignments ($36.7\%$ vs. $34.6\%$ without pre-training, $40.3\%$ vs. $38.7\%$ with pre-training). Reasons why learning joint-alignments does not help even more may include noise in the annotation data, making if difficult to learn exact relations, as well as a higher chance of overfitting to entire scenes as opposed to single alignments. When comparing results with and without synthetic pre-training we find significant improvement of 4\%. This indicates that even training on a different set of CAD models synthetically rendered in random poses provides useful training signals that transfer to real images. Inspecting Fig. \ref{fig_confidence}c we find that the pre-trained model achieves both a lower train and test loss leading to a higher instance alignment accuracy on the test set.\\
\textbf{Sparser and Faster.}
Another advantage of performing alignments for multiple CAD models jointly as opposed to in sequence is that it is a lot faster. The times in Tab. \ref{table_main_results} include the time for processing the input data (23 ms, for the main network architecture and inputs in row 4)) as well as a forward pass through the network (31 ms). These steps have to be repeated four times for the refinement procedure (3 refinement + 1 final classification score) from four different initialisations (see Fig. \ref{fig_confidence}a) leading to a total time of $4 \times 4 \times (23 + 31) = 864$ ms. By processing very sparse inputs i.e. $N_{bbox} = 200$ and $N_{CAD} = 200$, reducing the number of latents $N_{latent} = 40$ and encoding input information jointly, we can reduce both the time for processing the inputs (16 ms) as well as the forward pass (14 ms) and almost halve the total run-time to 480 ms. If not initialised from four different rotations (as would be realistic for example in a video setting where the rough object rotation is known from previous frames) this approaches the speed of single-shot methods while being considerably more accurate. Interestingly, this network variant is more accurate than the one encoding the full inputs jointly in the second row. This may indicate that it is easier for the network to learn to separate information for multiple alignments when presented with fewer inputs. Row 8 shows results for even sparser inputs, resulting in further small gains in speed.\\
\textbf{Learned classification score.}
Rather than just predicting pose updates we also learn classification scores indicating whether a given alignment is accurate or not. We use these learned classification scores to select the best alignment from multiple rotation initialisations (see Fig. \ref{fig_confidence}a) as well as to select from multiple predictions of the same object from different images in the Scan2CAD \cite{scan2cad} evaluation. We compare to the 2D detection confidence from ROCA \cite{roca} and note a small improvement (40.3\% compared to 38.8\%). More importantly, plotting the mean accuracy of the predictions sorted by the confidence we find that our 3D classification score is significantly better calibrated (see Fig. \ref{fig_confidence}b).\\
\textbf{Comparison to other methods - per-scene evaluation.}
We compare our method to other state-of-the-art CAD model alignment procedures \cite{total_3D_understanding,mask2cad,roca,sparc}. Quantitatively comparing against those methods we find that we improve significantly upon the instance alignment accuracy from $31.8\%$ to $40.3\%$ and the class mean accuracy from $24.9\%$ to $30.3\%$. We also improve in most categories with the notable exception of displays. Here our learned classification score struggles to distinguish between front and back-facing displays which look very similar when only sparse pixels are provided (see Fig. \ref{fig_qualitative} last column).\\
\textbf{Comparison to other methods - per-image evaluation.}
The advantages of our method compared to previous methods are even more pronounced on the per-image evaluation then they were on the per-scene evaluation (see Tab. \ref{table_per_image_results}). The class and instance alignment accuracy almost double compared to previous methods ($28.1\%$ vs. $16.1\%$ and $31.3\%$ vs. $18.4\%$). AP50 and APmean show even greater relative improvements, e.g. at $\rho=0.5$ AP50 improves from $10.8\%$ to $27.0\%$ and APmean improves from $3.0\%$ to $11.5\%$.
The reason why the improvements of our method compared to the previous ones are even more pronounced on the per-image compared to the per-scene evaluation is that the per-scene evaluation requires only one very accurate prediction for each object from any frame whereas the per-image evaluation has a high number of challenging viewpoints. Here both the multi-object predictions as well as the synthetic pre-training significantly increase the accuracy of the predictions.

\begin{table}[t]
    \centering
    \resizebox{\textwidth}{!}{
    \begin{tabular}{c|c|c|cc|cc|cc|c|cc}
     & \cellcolor[gray]{0.8} & \multicolumn{7}{c}{\textbf{$\text{AP}^{\text{mesh}}$}} & \cellcolor[gray]{0.8} & \multicolumn{2}{c}{\textbf{Alignment Accuracy}} \\
     \hline
     & \cellcolor[gray]{0.8} & &AP50 & APmean & AP50 & APmean & AP50 & APmean & \cellcolor[gray]{0.8} & class & instance \\
     \hline
      & \cellcolor[gray]{0.8} & $\rho$ & 0.3 & 0.3 & 0.5 & 0.5 & 0.7 & 0.7 & \cellcolor[gray]{0.8} & - & - \\
     \hline
     ROCA \cite{roca} & \cellcolor[gray]{0.8} & &1.8 & 0.4 & 10.8 & 3.0 & 20.3 & 7.1 & \cellcolor[gray]{0.8} & 16.1 & 18.4\\
     SPARC \cite{sparc} & \cellcolor[gray]{0.8} & & 2.4 & 0.5 & 9.8 & 3.0 & 19.1 & 7.0 & \cellcolor[gray]{0.8} & 15.9 & 17.4 \\
     Ours & \cellcolor[gray]{0.8} & & \textbf{11.6} & \textbf{3.4} & \textbf{27.0} & \textbf{11.5} & \textbf{36.4} & \textbf{18.7} & \cellcolor[gray]{0.8} & \textbf{28.1} & \textbf{31.3}\\
    \end{tabular}}
    \vspace{0.1cm}
    \caption{\textbf{Per-image alignment accuracy and $\text{AP}^{\text{mesh}}$ score on ScanNet \cite{scannet}}. Both AP scores and alignment accuracies are reported in \%. The $\rho$ value controls the threshold for computing the F1 score in the AP calculation. Smaller $\rho$ values require points sampled from the predicted aligned CAD model and the GT aligned CAD model to be closer together and therefore more accurate poses. Before computing the F1 score both CAD models are re-scaled isotropically such that the longest side of the 3D bounding box of the GT CAD model is equal to 10. Therefore for a typical object of maximum width and height equal to 1 m $\rho=0.5$ requires points sampled from the predicted CAD model to be within 5 cm of the GT CAD model and vice versa.}
    \label{table_per_image_results}
    \vspace{-0.4cm}
\end{table}

\section{Conclusion}
We introduced a novel render-and-compare approach that jointly aligns multiple CAD models to objects in an image. This provides advantages for both speed and accuracy at test time, improving the run-time by a factor of up to 5 and improving the instance alignment accuracy on ScanNet \cite{scannet} from $31.8\%$ to $40.3\%$. We demonstrate that some of this improvement stems from pre-training our network on a large number of random synthetic scenes. The fact that those scenes contain objects different to the ones the network is tested on highlights the ability of our render-and-compare approach to generalise.
Furthermore, we learn to predict not just pose updates but also classification scores that can be used for selecting a final pose from different candidates.
In the future we would like to extend render-and-compare to multi-view scenarios as well as using larger foundational models in a render-and-compare setting to reconstruct 3D scenes.

\newpage

\bibliography{egbib}

\newpage

\appendix

\section*{Supplementary Material}

Here we provide additional information to our main work. Sec. \ref{sec_training_depth_surface} provides information for the training of the surface normal and depth estimation networks. In Sec. \ref{sec_joint_training_pose_updates} we explain in detail what training examples are used for training Multi-SPARC-Net to predict pose updates and classification scores. Sec. \ref{sec_larger_number_CAD_models} provides ablation experiments for the maximum number of joint predictions and the size of the latent space. Sec. \ref{sec_artifacts} explains what CAD models were used for the synthetic pre-training. Finally, we highlight the major points of the provided video showing qualitative results on ScanNet in Sec. \ref{sec_video}. 
\section{Details for Training Surface Normal and Depth Networks}
\label{sec_training_depth_surface}

We utilize a lightweight convolutional encoder-decoder architecture \cite{GB-mono-2018-densedepth} to estimate both surface normals ($\mathbf{N}_{\text{Img}}$) and depth ($\mathbf{D}_{\text{Img}}$). The per-pixel probability distribution for each task is predicted, and the network is trained by minimizing the negative log-likelihood (NLL) of the ground truth. Learning parameters of probability distributions allows the networks to predict high uncertainty around object edges where GT annotations can often be wrong. This improves the quality of training \cite{GB-SNfromRGB_21_BAE}. The distribution for surface normals is parameterized using the Angular vonMF distribution proposed in \cite{GB-SNfromRGB_21_BAE}, while the depth distribution is parameterized with a Gaussian distribution. After training, we only consider the predicted mean values, discarding the uncertainty. The ground truth surface normals are provided by \cite{GB-SNfromRGB_19_FrameNet}, and the ground truth depth is obtained from ScanNet \cite{scannet}, following the train/test split. For depth estimation, we train the network on the available two million train images, while for surface normals, we train on the annotated images provided by \cite{GB-SNfromRGB_19_FrameNet} within the train image set, resulting in approximately 200K train images. Both networks are trained for ten epochs using the AdamW optimizer \cite{adamw}, and the learning rate is scheduled using the 1cycle policy \cite{cyclic_lr} with $lr_\text{max} = 3.5 \times 10^{-4}$ (same as \cite{Bae_2022_CVPR_depth}). A batch size of four is used for training both surface normals and depth networks. The steps described are consistent with the training protocol of \cite{sparc}.

\section{Joint Training Pose Updates and Pose Classifier}
\label{sec_joint_training_pose_updates}
We train Multi-SPARC-Net to predict both pose updates as well as a classification score $\sigma$ indicating whether the initial pose is correct. Similar to our evaluation a pose is classified to be correct if its translation $\Delta \textbf{T}$ is within $20$ cm, its rotation $\Delta \textbf{R}$ within $20^\circ$ and its scale $\Delta \textbf{S}$ within $20\%$ of the ground truth values. If we simply trained the network to predict classification scores from the initialisation used for learning pose updates, the classification scores would be heavily biased towards correct alignments as even after just a single CAD pose update the vast majority of poses would classify as correct. We therefore sample separate examples for training the pose classifier. For these separate examples no loss is backpropagated for the predicted pose updates (similar no loss is backpropagated from the classification score for the examples used to train the pose updates). Examples for training the pose classifier versus the pose updates are sampled with a ratio of 1:3. \\
\textbf{Sampling poses for training the pose classifier}.
\begin{table}[t]
    \centering
    \resizebox{\textwidth}{!}{
    \begin{tabular}{C{3cm}|C{4cm}C{2cm}C{2cm}C{3cm}|C{4cm}}
     Region & $\Delta \textbf{R}$ [tilt, azim, elev] in $[^\circ]$ & $\Delta \textbf{T}$ [cm] & $\Delta \textbf{S} [\%]$ & Discrete Rotation & Sampling Frequency\\
     \hline
    Region 1 & [7,10,10] & 13 & 13 & False & 0.4\\
    Region 2 & [7,10,10] & 13 & 13 & True & 0.2\\
    Region 3 & [7,45,20] & 30 & 30 & False & 0.2\\
    Region 4 & [7,45,20] & 30 & 30 & True & 0.1\\
    Region 5 & [20,45,20] & 60 & 60 & True & 0.1\\
    \hline
    \end{tabular}}
    \vspace{0.1cm}
    \caption{\textbf{Different pose regions for drawing samples when learning to classify poses.} $\Delta \textbf{R}$, $\Delta \textbf{T}$ and $\Delta \textbf{S}$ denote the maximum bounds from the ground truth values in which range rotation $\textbf{R}$, translation $\textbf{T}$ and scale $\textbf{S}$ are sampled uniformly. When discrete rotation is set to True the rotation is randomly rotated (with equal probability) by $0^\circ$, $90^\circ$, $180^\circ$ and $270^\circ$ around the vertical. The sampling frequencies denote the relative frequency for sampling from the different regions.}
    \label{table_regions}
\end{table}
Those examples are sampled from the five different regions stated in Tab. \ref{table_regions}. Here $\Delta \textbf{R}$, $\Delta \textbf{T}$ and $\Delta \textbf{S}$ denote the maximum bounds from the ground truth values in which range rotation $\textbf{R}$, translation $\textbf{T}$ and scale $\textbf{S}$ are sampled uniformly. Note that $\Delta \textbf{R}$ denotes the difference to the ground truth rotation in terms of Euler angles for tilt, azimuthal and elevation angle (in order). $\Delta \textbf{T}$ denotes the maximum difference to the ground truth translation in cm for $x$, $y$ and $z$ component. $\Delta \textbf{S}$ indicates the maximum difference in \% to the ground truth scale values in the three axes.
Discrete rotation implies that the rotation is randomly rotated (with equal probability) by $0^\circ$, $90^\circ$, $180^\circ$ and $270^\circ$ around the vertical. This is to ensure that the network is exposed to examples with the correct translation and scale but with a wrong rotation which it will encounter at test time as we use the classification score to select the final alignment from different rotation initialisation. Note that the symmetry of objects is taken into account when determining if a pose should be classified as correct or not. Poses from these regions are sampled with frequency as indicated in the last column in Tab. \ref{table_regions}. 
While the exact numbers in the sampling regions above do not matter, it is important to roughly balance the number of correct and incorrect poses and to ensure that the poses the network is likely to encounter at test time are covered in the training examples.\\
\textbf{Sampling poses for training the pose updates}.
For learning the pose updates we sample the initial pose as follows. $\textbf{T}$ is sampled by uniformly sampling a point within the predicted bounding box and then lifting that point into 3D by providing a $z$ value sampled $z \sim \text{Uniform}(1,5) $ in metres. The scale $\textbf{S}$ is sampled uniformly within the range of the minimum object scale and maximum object scale for all CAD model alignments of this category on ScanNet \cite{scannet} by \cite{scan2cad}. Finally, $\textbf{R}$ is sampled uniformly within $10^\circ$ tilt, $45^\circ$ azimuthal and $20^\circ$ elevation angle of the ground truth rotation.

\section{Simultaneous Prediction for Larger Number of CAD models}
\label{sec_larger_number_CAD_models}
In Tab. \ref{table_ablation} we show results when varying the maximum number of joint predictions $N_\text{mul}$ and the number of latents $N_\text{latent}$ dedicated to each alignment. Here we observe that compared to our main setup (highlighted in yellow) we can reduce the number of latents by a factor of two while still achieving very similar accuracies. Further, we note that increasing the number of joint alignments also achieves similar results. This brings extra advantages for speed when reconstructing scenes containing many objects.

\begin{table}[t]
    \centering
    \resizebox{\textwidth}{!}{
    \begin{tabular}{c|ccccccccc|cc|c}
     Method & bathtub & bed & bin & bkshlf & cabinet & chair & display & sofa & table & \cellcolor[gray]{0.8} \textbf{class} & \cellcolor[gray]{0.8} \textbf{instance} & time [ms] \\ [0.5ex] 
     \hline
     Number of Instances \# & 120 & 70 & 232 & 212 & 260 & 1093 & 191 & 113 & 553 & \cellcolor[gray]{0.8} 9 & \cellcolor[gray]{0.8} 2844 & - \\
     \hline
     \hline
    $N_{\text{mul}} = 5$ , $N_{\text{latent}} = 40$ & 20.0 & 24.3 & 41.4 & 12.7 & 16.2 & 60.2 & 3.7 & 28.3 & 23.7 & \cellcolor[gray]{0.8} 25.6 & \cellcolor[gray]{0.8} 36.4 & 848\\
    \cellcolor{yellow} $N_{\text{mul}} = 5$ , $N_{\text{latent}} = 80$ & \cellcolor{yellow} 20.0 & \cellcolor{yellow} 28.6 & \cellcolor{yellow} 40.1 & \cellcolor{yellow} 13.7 & \cellcolor{yellow} 20.4 & \cellcolor{yellow} 59.9 & \cellcolor{yellow} 0.5 & \cellcolor{yellow} 36.3 & \cellcolor{yellow} 23.0 & \cellcolor{yellow} 26.9 & \cellcolor{yellow} 36.7 & \cellcolor{yellow} 864 \\
    $N_{\text{mul}} = 10$ , $N_{\text{latent}} = 20$ & 20.0 & 25.7 & 40.1 & 11.8 & 20.4 & 60.7 & 4.7 & 29.2 & 19.5 & \cellcolor[gray]{0.8} 25.8 & \cellcolor[gray]{0.8} 36.1 & 944\\
    $N_{\text{mul}} = 10$ , $N_{\text{latent}} = 40$ & 16.7 & 27.1 & 41.8 & 13.7 & 18.8 & 57.8 & 2.6 & 31.9 & 21.5 & \cellcolor[gray]{0.8} 25.8 & \cellcolor[gray]{0.8} 35.4 & 960\\
    \hline
    \end{tabular}}
    \vspace{0.1cm}
    \caption{\textbf{Ablation.} Alignment Accuracy on ScanNet      
    \cite{scannet,scan2cad} when varying the number of CAD models $N_\text{mul}$ for which predictions are made jointly and the number of latents $N_\text{latent}$ dedicated to the processing of the information for each alignment. Times are for reconstructing a scene containing containing 5 or 10 objects respectively. Note that doubling the number of objects only leads to a marginal increase in time. The yellow row highlights the main setup for which no pre-training was performed from the results table in the paper.}
    \label{table_ablation}
    
\end{table}

\section{Artifacts Rendering ShapeNet Normals}
\label{sec_artifacts}
\begin{figure*}[t]
    \centering
    \includegraphics[width=1.0\linewidth]{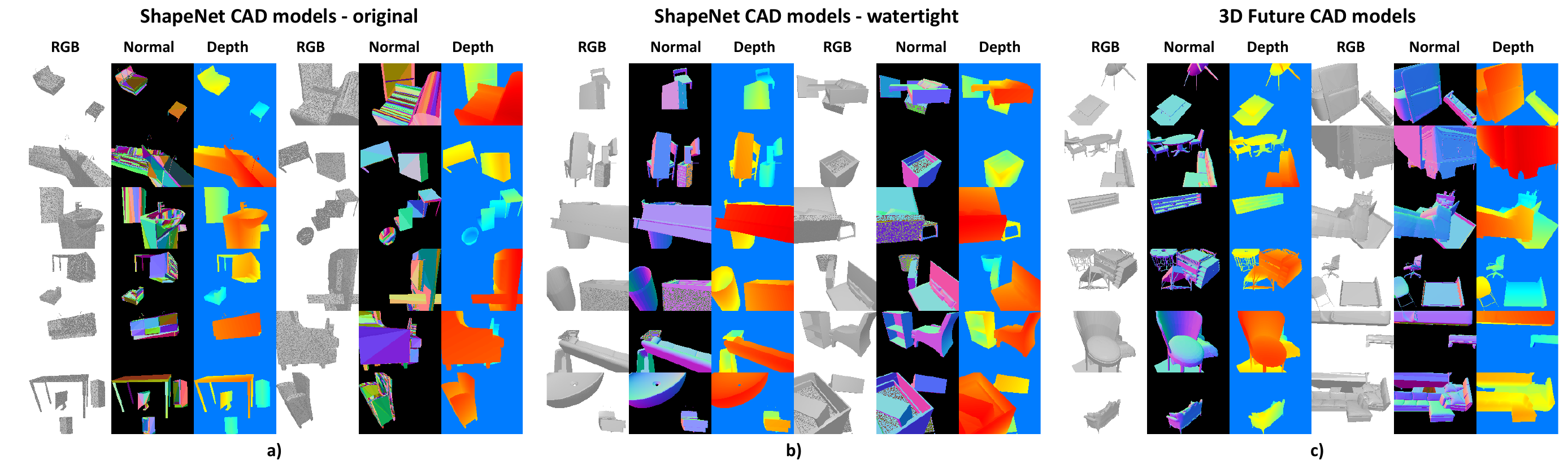}
    \vspace{-0.3cm}
    \caption{\textbf{Comparison when rendering ShapeNet \cite{chang2015shapenet} CAD models and 3D Future \cite{3d_future} CAD models using PyTorch3d \cite{pytorch3d}.} We find that rendering normals for the original (a) as well as the processed watertight ShapeNet \cite{chang2015shapenet} CAD models (b) results in artifacts and therefore use CAD models from 3D Future \cite{3d_future} instead which result in correct renders (c).}
    \vspace{-0.1cm}
    \label{fig_surface_normals}
\end{figure*}

For the synthetic pre-training we render CAD models in a large number of random poses and train Multi-SPARC-Net on those poses. Objects in the scenes from ScanNet \cite{scannet} were annotated by \cite{scan2cad} with CAD models from ShapeNet \cite{chang2015shapenet}. Those CAD models are used for the main training and evaluation. However, when rendering ShapeNet \cite{chang2015shapenet} CAD models with PyTorch3D \cite{pytorch3d} we get the semi-random patterns for surface normals as seen in Fig. \ref{fig_surface_normals}a. This is because the simple Normal Shader we implemented in PyTorch3D \cite{pytorch3d} identifies which CAD model face is rendered into a given pixel and then interpolates the per-vertex surface normal value from the three vertices of the given face. The issue with rendering the original ShapeNet \cite{chang2015shapenet} models is that many of them are not closed or contain doubly-meshed faces, meaning that the mesh contains two identical faces with the order of two vertices swapped such that their surface normals will point in opposite directions. Rendering these means that front or back-facing surface normals will be selected at random for the interpolation, leading to the wrong surface normal renderings in Fig. \ref{fig_surface_normals}a. We try to make ShapeNet \cite{chang2015shapenet} CAD models watertight with consistently oriented faces using \cite{huang2020manifoldplus}. While this allows us now to render CAD normals (Fig \ref{fig_surface_normals}b) the generated surface normal renders sometimes still suffer from noise when the procedure for making the CAD models watertight did not succeed. Further, making the models watertight using \cite{huang2020manifoldplus} increases the median number of vertices from just 600 to 130 K which makes the CAD models impractical for us to use as it massively increases the rendering time. Instead we choose to perform our synthetic training on CAD models from 3D Future \cite{3d_future} which are already watertight and yield correct surface normals when rendering in PyTorch3D \cite{pytorch3d} (see Fig. \ref{fig_surface_normals}c).

\section{Video with Qualitative Visualisation}
\label{sec_video}
\begin{figure*}[t]
    \centering
    \includegraphics[width=1.0\linewidth]{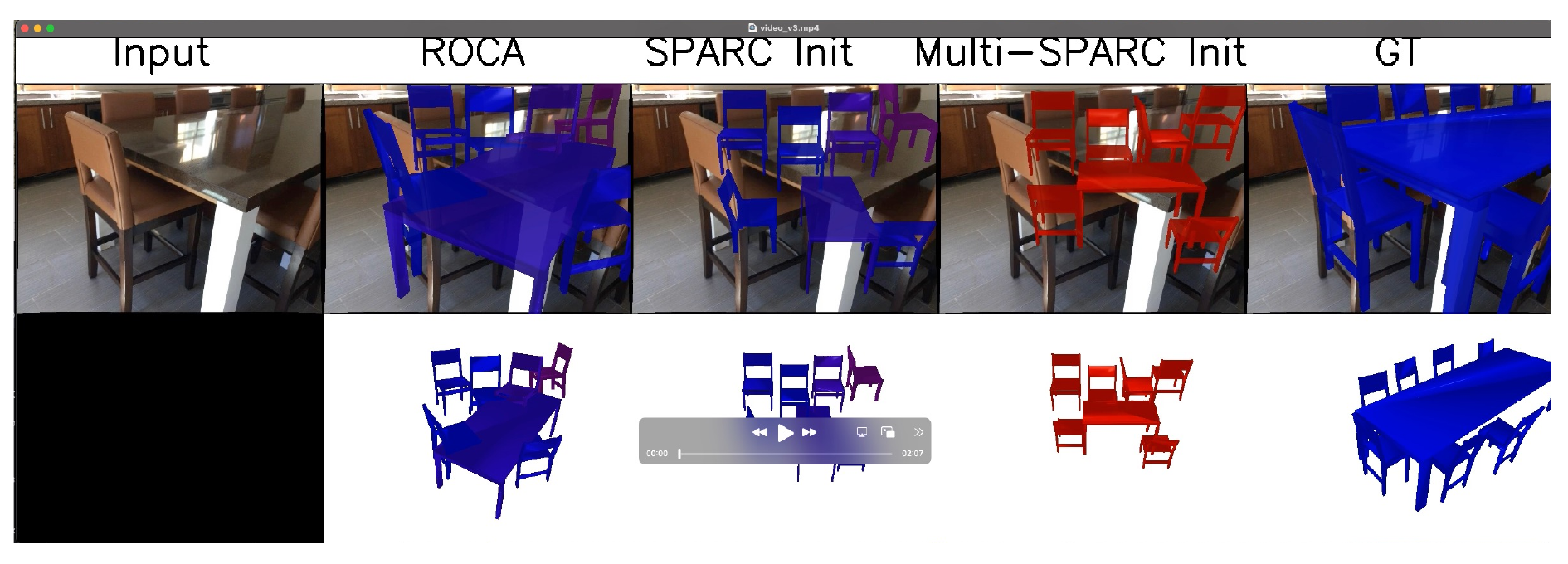}
    \vspace{-0.3cm}
    \caption{\textbf{Video with Qualitative Visualisations.}}
    \vspace{-0.3cm}
\end{figure*}

We provide a video comparing our predictions qualitatively to ROCA \cite{roca} and SPARC \cite{sparc} on ScanNet \cite{scannet} (\url{https://www.youtube.com/watch?v=NtOU5BOmagw}). Note that for our predictions the color indicates the value of the learned classification score. For ROCA \cite{roca} and SPARC \cite{sparc} the color indicates the value of the 2D detection score which is not updated based on the refinement for SPARC \cite{sparc}. As explained in Sec. 3.3 of the main paper we initialise CAD models from four different rotations. In the video we only show those alignments corresponding to that rotation initialisation which led to the highest classification score after three refinements. This is why some of the initialisation between SPARC \cite{sparc} and ours are different. However, apart from selecting the rotation initialisation, the initialisations between SPARC \cite{sparc} and ours are the same.\\
While ROCA's \cite{roca} predictions are often inaccurate, suffering from displacements and wrong scale predictions, we find that our predictions are accurate. Particularly, they are are also more accurate than SPARC's \cite{sparc} predictions, matching object outlines more closely due to the synthetic pretraining and yielding more accurate alignments for objects in close proximity to each other due to the mulit-object training and predictions. Interestingly, we find that Multi-SPARC-Net learns to rotate objects by more than $90^\circ$ around the vertical over multiple refinements even though at train time it only ever learned to predict updates that were less than $45^\circ$ away from the correct rotation.

\end{document}